\documentclass{article}

     \usepackage[final]{neurips_2020}

\usepackage[utf8]{inputenc} 
\usepackage[T1]{fontenc}    
\usepackage{hyperref}       
\usepackage{url}            
\usepackage{booktabs}       
\usepackage{amsfonts}       
\usepackage{nicefrac}       
\usepackage{microtype}      
\usepackage{graphicx}
\usepackage{listings}
\usepackage{float}
\usepackage{amssymb}
\usepackage{siunitx}
\usepackage{amsmath}
\usepackage[english]{babel}
\setcitestyle{authoryear,open={[},close={]}} 

\title{End-to-End Attention-based Image Captioning}

\author{%
  Carola Sundaramoorthy\\
  A0225543J \\
  \texttt{e0576177@u.nus.edu} \\
  \And
  Lin Ziwen Kelvin \\
  A0150057N \\
  \texttt{e0014961@u.nus.edu} \\
  \And
  Mahak Sarin \\
  A0225535H \\
  \texttt{e0576169@u.nus.edu} \\
  \And
  Shubham Gupta \\
  A0225160U \\
  \texttt{e0575794@u.nus.edu} \\
}

\begin{document}

\maketitle

\begin{abstract}
In this paper, we address the problem of image captioning specifically for molecular translation where the result would be a predicted chemical notation in InChI format for a given molecular structure. Current approaches mainly follow rule-based or CNN+RNN based methodology. However, they seem to underperform on noisy images and images with small number of distinguishable features. To overcome this, we propose an end-to-end transformer model. When compared to attention-based techniques, our proposed model outperforms on molecular datasets.

\end{abstract}

\section{Introduction}

Molecular representation has remained an interest of scientific curiosity since the 19\textsuperscript{th} century \cite{19cen}. Tradition representations are described pictorially using bond and atom notations. These representations are not very useful for the computational processing of chemical structures. Several linear notations have been developed for describing a molecule. Hill notation is one of the earliest such notations. For bromoethane, the notation would look like C\textsubscript{2}H\textsubscript{5}Br. This notation only gives the information about the atoms involved but does not provide any insight into how the atoms are linked together.

\begin{figure}[htbp]
    \centerline{
    \includegraphics[scale=0.5,width=0.6\columnwidth]{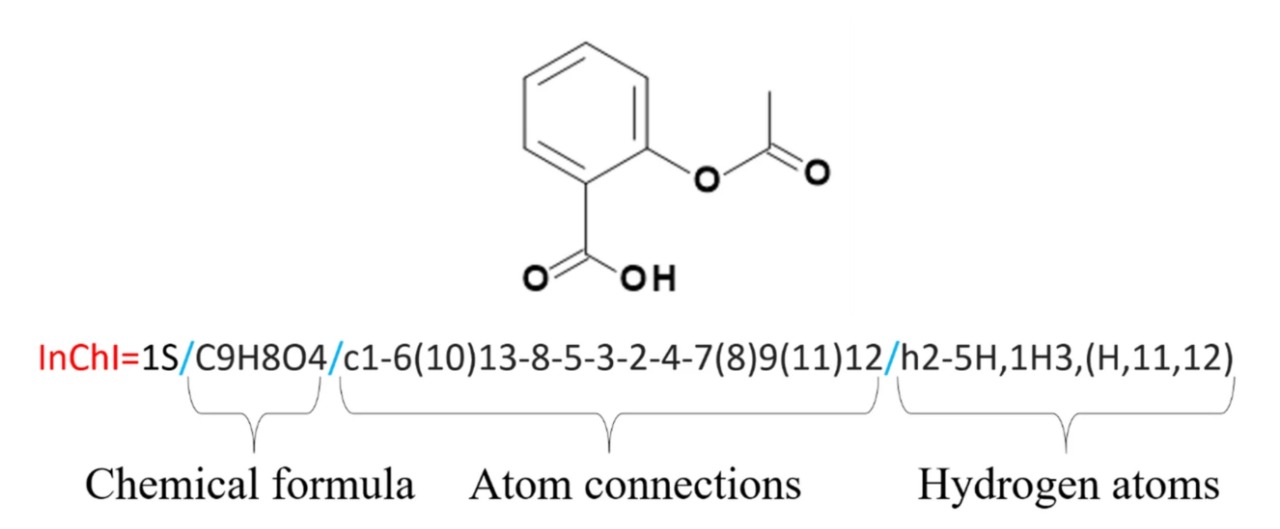}}
    \caption{InChI notation for the shown molecule}
    \label{fig:molecularExample}
\end{figure}

One of the more detailed and computer-readable notations is InChI or International Chemical Identifier developed by the International Union of Pure and Applied Chemistry \cite{inchi}. InChI describes molecules with layers of information, such as the Main, Charge, Stereochemical, and Isotopic layers, which are further constituted of sublayers. The Chemical formula, Atom connections, and Hydrogen atoms sublayers forms the main layer as shown in Figure ~\ref{fig:molecularExample}\cite{inchiguide}.In this work, we will be using the InChI notation. 

Often in drug discovery projects, there is a need to analyse and take advantage of the published experimental data. However most of the publications do not include computer-readable notations such as InChI, SMILES etc, instead, they contain structural diagrams similar to that in  Figure ~\ref{fig:molecularExample}. There exist segmentation methods to extract these images from the publications \cite{segment}. There are also several tools to predict the chemical properties such as solubility, toxicity etc from the computer-readable notations \cite{chemixnet, smiles2vec}. A tool to convert these images to computer-readable notations will be useful for many such applications.

The goal of image captioning is to automatically generate InChI descriptions for a given image, i.e., to capture the relationship between the different shapes and molecular names present in the image, generate the expressions. While present techniques to solve the image captioning task revolves around using image segmentation approaches such as SETR \cite{imageseg} or some rule-based feature extraction for performing the molecular translation task but they under-perform when the input molecular image is corrupted with noise, which is often the case due to artefacts from scanning old journals. 
Motivated by the recent success of the transformer architecture in both computer vision and natural language processing task, we propose to use complete transformers based model for this molecular translation.

\section{Literature Review}

Image captioning is a challenging multimodal task and has been a well studied task with \cite{hossain2019comprehensive} being a comprehensive survey. As a non-deep learning approach, \cite{rennie2017self} makes use of policy gradient with reinforcement learning to generate the captions for an image. While deep learning methods are largely based on encoder-decoder architecture where the encoders are used to as a mechanism to extract features and decoders are used to predict words according to the feature extracted by the encoder network \cite{mao2014deep,vinyals2016show}.

Show, Attend and Tell model \cite{xu2015show} introduced the encoder-decoder architecture and the visual attention mechanism for image captioning in a simple yet powerful approach. This method performs relatively well for captioning, during the caption generation, the decoder only considers the input image in its entirety. Attention mechanisms allow the decoder to focus on specific parts of the image to generate more accurate captions. Since this model dynamically selects image features on which the attention mechanism works are obtained from a deep convolutions encoder that misses out on low-level specific details \cite{topBottom}. In this way, features may often fail to attend and provide the direct cues i.e. specific details to the decoder which fail to generate the next caption properly.

Following the success of Transformer in the field of natural language processing \cite{attention} uses transformers in the decoder instead of RNN which results in an excellent performance. However, the encoder still has a CNN model such as ResNet \cite{resnet} which is pre-trained on image classification task to extract spatial feature or to use a Faster-RCNN \cite{fasterrcnn} which is pre-trained on object detection task to extract bottom-up feature.

ViT \cite{vit} model uses a transformer for image classification making it convolution-free operation and shows promising performance mostly when it is pre-trained on very huge datasets. Since then, convolution-free methods are used for both high-level and low-level downstream tasks \cite{topBottom}. 

In the context of molecular translation, the usage of feature extraction with hand-written rules to identify chemical groups was proposed by \cite{rule1, sadawi2012chemical}, while \cite{park2009automated} made use of image segmentation. These methods are highly complicated and very difficult to improve upon. Moreover, they also require significant domain expertise and are very time consuming to develop. In \cite{chemception}, Garrett \emph{et al.} showed that the deep learning approach with minimal chemistry knowledge outperforms the rule-based expert developed models. Deep learning with attention models were also proposed \cite{staker2019molecular, decimer}, making use of a CNN and RNN with attention. Other methods include optical graph recognition of molecular compounds \cite{chemgrapher}.

We propose a complete transformers based model where Vit \cite{vit} transformer network would be used to replace the CNN in the encoder part. This model is more simple yet effective as it avoids convolution operation at the encoder stage, thus enabling it to utilize long-range dependencies among the patches from the very beginning via the self-attention mechanism. We evaluate our method on the molecular dataset and it outperforms models based on the encoder-decoder structure of CNN+RNN or CNN+Transformer. We used the molecular dataset for evaluating our model as this dataset includes noisy data and the images have less distinguishable features compared to the MSCOCO dataset.

\section{Framework}

\begin{figure}[H]
    \centering
    \includegraphics[width=0.9\textwidth]{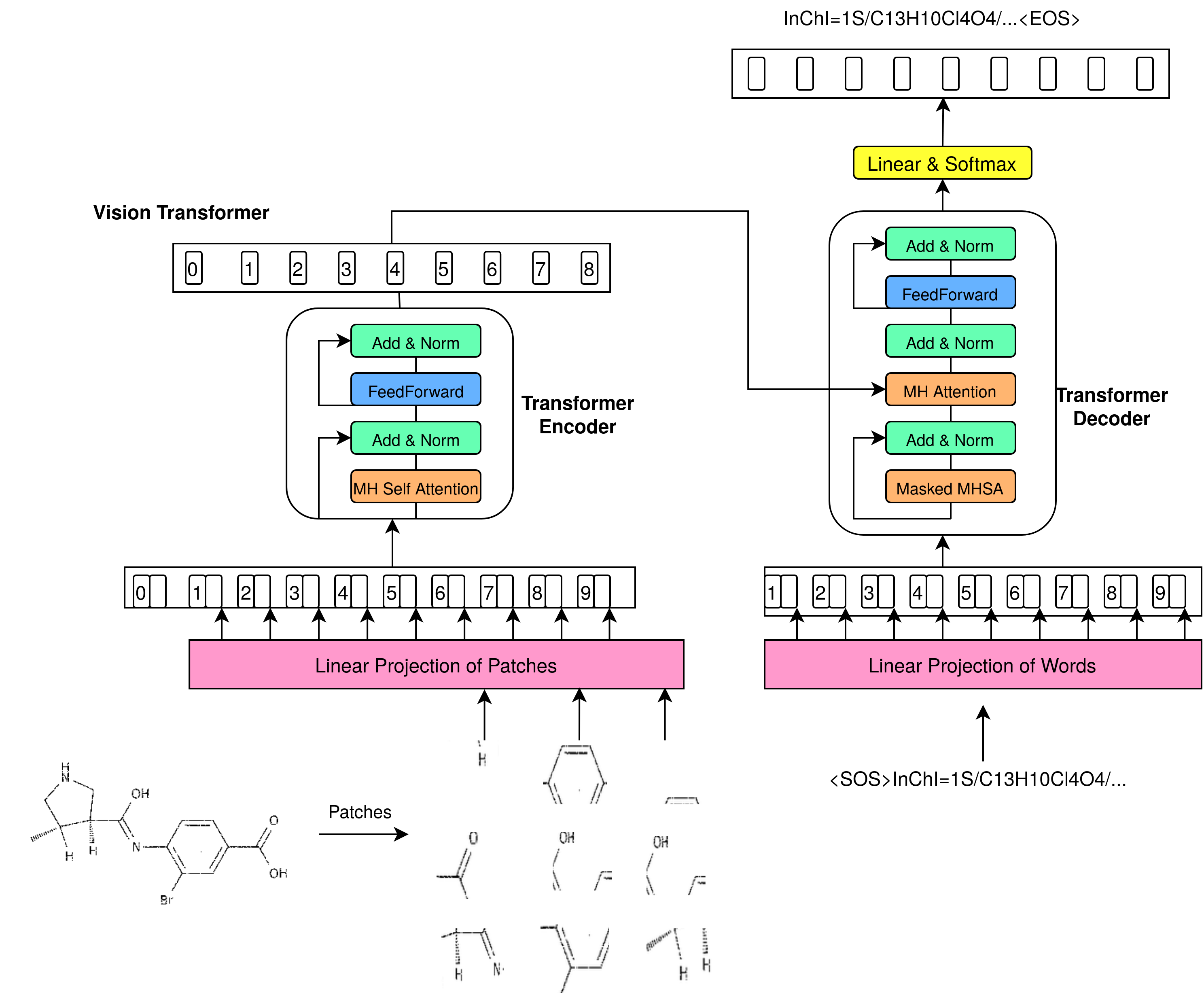}
    \caption{Architecture Diagram}
    \label{fig:final_attention_architecture}
\end{figure}

Our approach consists of a Vision Transformer as the encoder and a Vanilla Transformer decoder. This model is trained end-to-end on 70GB GPU system. We have implemented it using PyTorch and PyTorch Lightning. The architecture is shown in Figure ~\ref{fig:final_attention_architecture} and consists of the following parts.

\subsection{Encoder-Decoder Model}

\subsubsection{Positional Encoding}
As our model is convolution free, to utilize the order of the sequence, it must use the information about the relative or absolute position of the tokens in the sequence we add "positional encoding". These encodings have the same dimensions so that it is additive. There are many choices of positional encoding, learned and fixed. In this paper, we use sine and cosine functions of different frequencies:

\begin{equation}
    \mli{PE}_{(pos,2i)} = sin(pos/1000^{2i/d_{model}})
\end{equation}
\begin{equation}
    \mli{PE}_{(pos,2i+1)} = cos(pos/1000^{2i/d_{model}})
\end{equation}

where pos is the position and i is the dimension. Each dimension of the positional encoding corresponds to a sinusoid. The wavelengths form a geometric progression from $2\pi$ to $10000.2\pi$. We chose this method because we hypothesized it would allow the model learn easily for any fixed offset k, since $\mli{PE}_{(pos+k)}$ it can be represented as a linear function of $\mli{PE}_{(pos)}$. \cite{attention}

\subsubsection{Encoder: Vision Transformer}

The standard Transformer receives a flat array sequence of token embeddings as input.
Since the input to our transformer are 2D images, we reshape the image ${x} \in \mathbb{R}^{H \times W \times C}$ into a sequence of flattened 2D patches ${x}_p \in \mathbb{R}^{N \times (P^2 \cdot C)}$, where $(H, W)$ is the resolution of the original image, $C$ is the number of channels, $(P,P)$ is the resolution of each image patch, and $N=HW/P^2$ is the resulting number of patches, which also serves as the effective input sequence length for the Transformer.
The transformer uses constant latent vector size $D$ through all of its layers, so we flatten the patches and map to $D$ dimensions with a trainable linear projection. We refer to the output of this projection as the Linear Projection of Patches. \cite{vit}.

Similar to BERT's \verb|[class]| token, we prepend a learnable embedding to the sequence of embedded patches (${z}_0^0={x}_\text{class}$), whose state at the output of the Transformer encoder (${z}^0_L$) serves as the image representation ${y}$.

Position embeddings are used along with the patch embeddings to retain positional information. We use standard learnable 1D position embeddings since we have not observed significant performance gains from using more advanced 2D-aware position embeddings. The resulting sequence of embedding vectors serves as input to the encoder.

The Transformer encoder consists of alternating layers of multiheaded(MHA) self-attention and MLP blocks. MHA consists of multiple heads $h_{i}$, each of which performs scaled dot product attention function, which allows the model to learn about different spaces. Then a linear transformation $W_0$ is used to aggregate the attention results of different heads, the process can be formulated as follows:
\begin{equation}
    \mli{MHA}(Q,K,V) = Concat(h_1...h_H)W^0
\end{equation}

The scaled dot product attention is computed using the formula:
\begin{equation}
    Attention(Q,K,V) = Softmax(\frac{QK^T}{\sqrt{d}_k})V
\end{equation}

where Q, K and V are the Query, Key and Value matrices.

Layernorm (LN) is applied before every block, and residual connections after every block.The Feed forward network contains two layers with a GELU non-linearity.

\begin{equation}
    \mli{FFN}(x) = \mli{FC}2(Dropout(\mli{GELU}(\mli{FC}1(x))))
\end{equation}
In each sublayer, there exists a residual connection which is followed by a layer normalization:
\begin{equation}
    x^{out} = LayerNorm(x^{in}+ Sublayer(x^{in}))
\end{equation}
where $ x^{in}$, $ x^{out}$ are the input and output of one attention/feed forward layer sublayer respectively.

\subsubsection{Decoder: Vanilla Transformer}
As input to the decoder, we first tokenize the InChI string of the image. The vocabulary consists of 275 tokens, including the special tokens <SOS>, <PAD> and <EOS>. After tokenization, we add sinusoid positional embedding to the word embedding features and take both the addition results and encoder output features as the input. 

The decoder consists of stacked layers with each layer consisting of a Masked Multi-Head Self Attention sublayer followed by a multi-head cross attention sublayer and a positional feed-forward sublayer sequentially.

During training, the output of the decoder is used to project the probabilities of a given token in the vocabulary via a linear layer. For the ground truth sentence $y_{1:t-1}^{*}$ and the prediction $y_{t}^{*}$ of captioning model $\theta$, we want to minimize the following cross-entropy loss:

\begin{equation}
    L_{XE} = - \sum_{t=1}^{T} log(p_{\theta}(y_{t}^{*}|y_{1:t-1}^{*}))
\end{equation}

During inference, we make use of auto-regressive decoding to predict the next letter of the molecule's InChI from the previous step and generate the molecule's structure.

\section{Experiment}

\subsection{Datasets}

For training the transformer models, we require a large dataset. So, we used the combined data from the following datasets:

\subsubsection{Bristol Myers Squibb}
Bristol-Myers Squibb is a global biopharmaceutical company working to transform patients' lives through science. They have provided access to a large set of synthetic image data consisting of more than 2.4 million images with the corresponding InChI format. 

\subsubsection{SMILES}
\textbf{SMILES} stands for Simplified Molecular Input Line Entry System.It is a specification in the form of a line notation for describing the structure of chemical species using short ASCII strings. 
We convert the SMILES strings into InChI format using the rdkit tool. Furthermore, we also generate the images for these images using the same tool.
The dataset was obtained from a \href{https://www.kaggle.com/c/smiles/data}{Kaggle contest}. 

\subsubsection{GDB-13}
GDB-13 enumerates small organic molecules containing up to 13 atoms of C, N, O, S, and Cl following simple chemical stability and synthetic feasibility rules. With approx. 977 million structures, GDB-13 is the largest publicly available small organic molecule database upto date. GDB-13 also provides data in the SMILES format. We use rdkit to convert the data to the required format. The dataset was obtained from the \href{https://gdb.unibe.ch/downloads/}{official website}.

\subsubsection{Augmented Images}
\begin{figure}[H]
    \centering
    \includegraphics[width=0.7\textwidth]{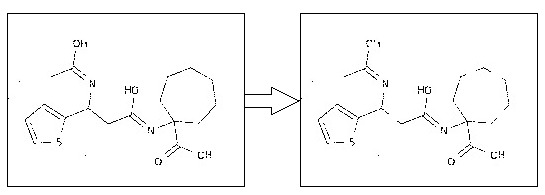}
    \caption{Sample augmented input}
    \label{fig:aug}
\end{figure}

We generate another dataset consisting of synthetic images. For these images, we use rdkit to modify the images for a given InChI string.
The modifications include salt and pepper noise, dropping an atom, converting double bonds to single bonds and introduce new artifacts as shown in Figure ~\ref{fig:aug}.
These noisy images help us create a generalized and robust model.

\subsection{Implementation Details}

The above-mentioned dataset is randomly split into training, test and validation sets containing 70\%, 10\% and 20\% images respectively. Input images are resized to 384*384 and patch size is set to 16. The feature dimension is 512. Both encoder and decoder contain 12 layers each and 12 attention head number. The whole model is first trained with cross-entropy loss for 10 epochs using an initial learning rate of \num{3e-5} and decayed by 0.5 during the last two epochs. We use Adam optimizer and the batch size is varied in [64-512]. Like other captioning methods, we also finetune our model using self-critical training \cite{train}.

\begin{table}[H]
  \caption{Hyperparameters of Model}
  \label{tab:hyperparam}
  \centering
  \begin{tabular}{lll}
    \toprule
    \cmidrule(r){1-2}
    Parameter     & Value     \\
    \midrule
    Image Size & 224x224 or 384x384 \\
    Num patches & 14 or 24 \\
    Patch Size & 16x16 \\
    Optimizer & Adam \\
    Initial Learning Rate & \num{3e-5} \\
    Batch Size & 64 \\
    \bottomrule
  \end{tabular}
\end{table}

\subsection{Optimizations for Inference Speed}

\subsubsection{JIT Compilation}
PyTorch JIT compiler optimizes PyTorch programs. It is a lightweight threadsafe interpreter, supports custom transformations and is not just for inference as it has auto diff support. TorchScript is designed specifically for obtaining high-performance in ML applications. It supports complex control flows, common data structures and user-defined classes. TorchScript creates an Intermediate Representation(IR) of the PyTorch models which can be compiled optimally at runtime by PyTorch JIT. During inference on GPU, we noticed a speed increase of 1.2-1.5x.

\subsubsection{Segregate Encoder and Decoder Modules}
During inference, we use autoregressive decoding to get the final molecule structure. The Transformer class provided in Pytorch is not optimized for speed. From the architecture diagram, we notice that the encoder output can be computed separately from the decoder. This means that the encoder output can be computed once and re-used for each timestep after. By default, Pytorch does not facilitate this and recomputes encoder output for each decoding step. Hence, we propose to separate the Encoder and Decoder modules.

\begin{figure}[H]
    \centering
    \includegraphics[width=0.4\textwidth]{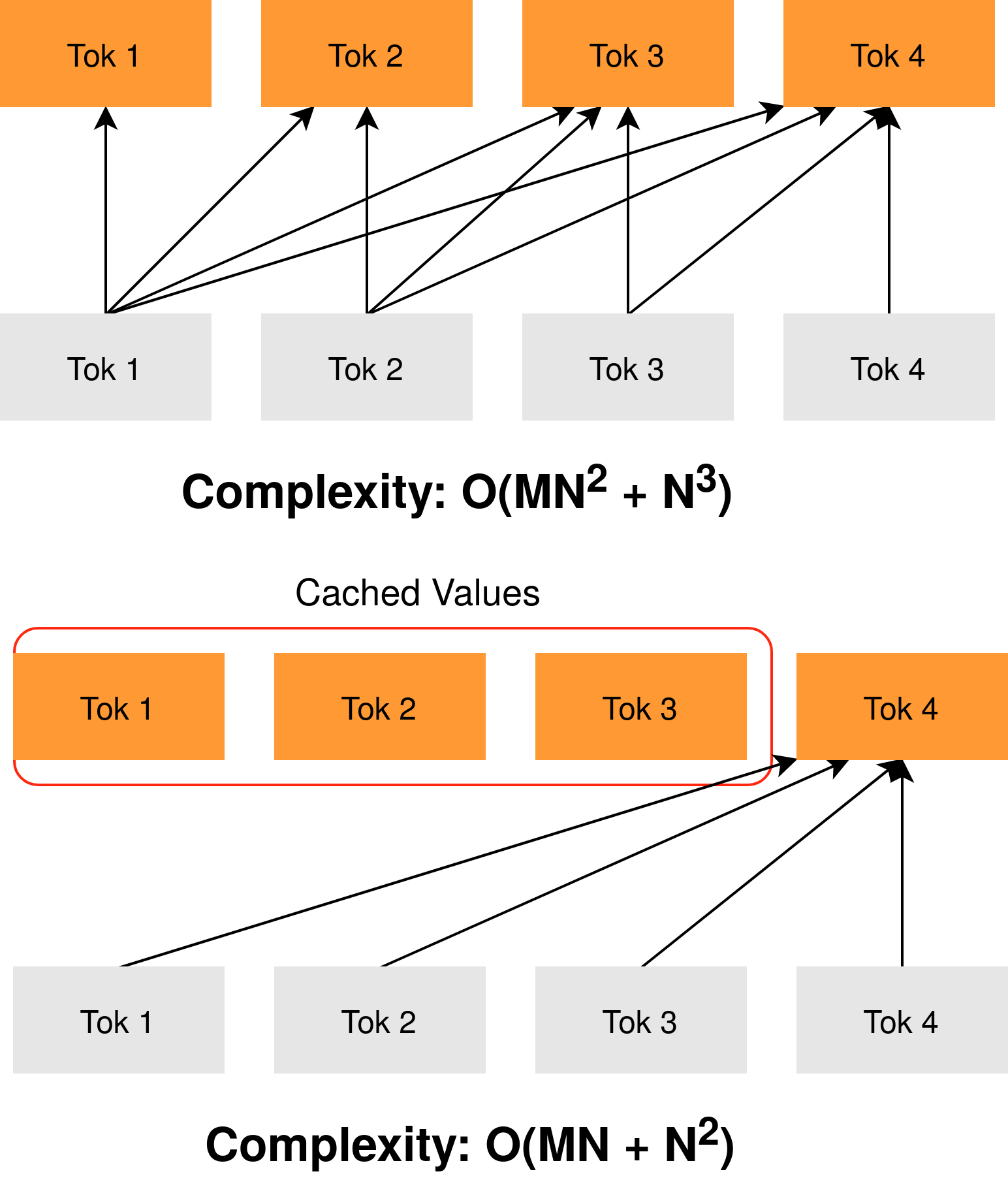}
    \caption{Caching Trick}
    \label{fig:encoder_optimization}
\end{figure}

\subsubsection{Reduce Time Complexity}
Building on the speed increase we obtained by segregating the encoder and decoder modules, we noticed that the embedding of a decoded token only depends on the tokens that were decoded before it. Hence, it is unnecessary to recompute the embeddings of the already decoded tokens and we can cache them instead. We choose to cache the output of each layer, instead of caching the query, key and value matrices. 
Computing the running time complexity, for the n-th output token, we know that performing self-attention is a $O(n^2)$. Furthermore, we also know that the encoder-decoder attention of dimension $M$ is a $O(Mn)$ operation. For a total of $N$ tokens and timesteps, we get the overall complexity as $O(MN^2 + N^3)$.

Through our caching trick, as shown in Figure ~\ref{fig:encoder_optimization}, we can speed up the computation at each time step. Now, only small parts of the self-attention and encoder-decoder attention mechanism are responsible for updating the last token. The new complexity per timestep reduces to $O(M+N)$, and therefore for $N$ timesteps, it will be $O(MN + N^2)$. 

\section{Results}
The training dataset consisted of the images from Bristol-Myers Squibb dataset ,SMILES,GDB-13 and augmented images. The performance of both the transformer-based and standard CNN+RNN modules are compared after training on the same dataset using cross-entropy loss function. Our final model was trained for 10 epochs with the hyper-parameters in Table \ref{tab:hyperparam}.

Various experiments were conducted for different number of layers in the encoder/decoder. To evaluate the performance of our model, we use the Levenshtein distance which measures the number of edits required to convert one string to another. The results are summarized in Table ~\ref{tab:resultTable}. Our model performs better with average Levenshtein distance of 6.95, compared to 7.49 of the ResNet/LSTM model with attention. We have also observed that having a large number of layers improves the accuracy of our model, where it becomes competitive with current approaches at 24 layers. 

\begin{table}[H]
  \caption{Levenshtein Distance of Different Models}
  \label{tab:resultTable}
  \centering
  \begin{tabular}{lllll}
    \toprule
    \cmidrule(r){1-2}
    Model   &Res & Layer & Epochs & Levenshtein Distance     \\
    \midrule
    Standard CNN+RNN &224 &3 & 10 & 103.7 \\
    Resnet18 + LSTM &224 &4 & 10 & 75.03 \\
    Resnet34 + LSTM &224 &4 & 10 &45.72 \\
    Resnet50 + LSTM &224 &5 & 10 & 7.49 \\
    ViT Transformers  & 224 & 3 & 5 &79.82 \\
    ViT Transformers & 224 & 6 & 5 &54.58 \\
    ViT Transformers  & 224 & 12 & 5 &31.30 \\
    ViT Transformers  & 384 & 24 & 10 &\textbf{6.95} \\
    \bottomrule
  \end{tabular}
\end{table}

\section{Conclusion}

We have shown that transformer-based models outperform present deep learning models in the task of molecular translation. The shallow layers in our model exploit both the local and global contexts by using different attention heads which can't be achieved by the CNN encoders. The middle layer pays attention to the primary molecular structure whereas the last layer pays attention to the complete molecular structure present in the image, capturing global context information at every encoder.

Since our model performed reasonably well on a dataset consisting of images with low distinguishable features, it is assured that it will perform well on MSCOCO dataset.

\section{Future Work}

Through our experiments, we found that increasing the image dimensions from 224x224 to 384x384 image and 16x16 patch size, has a large positive impact on the model performance. This is expected since the length of the patch increases from 196 to 576. Since the vanilla transformer has a max token length of 512, we can explore using architectures such as the LongFormer\cite{beltagy2020longformer} to deal with these longer sequences. 

To improve training time, we can implement the Layer-wise Adaptive Moments optimizer for Batch training[LAMB]\cite{LAMB} optimizer, which uses principled layerwise adaptation strategy to accelerate training of deep neural networks using large mini-batches. This can be implemented by training the model using the Microsoft DeepSpeed framework.

\clearpage
\bibliography{main}

\begin{thebibliography}{27}
\providecommand{\natexlab}[1]{#1}
\providecommand{\url}[1]{\texttt{#1}}
\expandafter\ifx\csname urlstyle\endcsname\relax
  \providecommand{\doi}[1]{doi: #1}\else
  \providecommand{\doi}{doi: \begingroup \urlstyle{rm}\Url}\fi

\bibitem[Anderson et~al.(2018)Anderson, He, Buehler, Teney, Johnson, Gould, and
  Zhang]{topBottom}
Peter Anderson, Xiaodong He, Chris Buehler, Damien Teney, Mark Johnson, Stephen
  Gould, and Lei Zhang.
\newblock “bottom-up and top-down attention for image captioning and visual
  question answering".
\newblock \emph{arXiv:1707.07998}, 2018.

\bibitem[Beltagy et~al.(2020)Beltagy, Peters, and Cohan]{beltagy2020longformer}
Iz~Beltagy, Matthew~E Peters, and Arman Cohan.
\newblock Longformer: The long-document transformer.
\newblock \emph{arXiv preprint arXiv:2004.05150}, 2020.

\bibitem[David et~al.(2020)David, Thakkar, Mercado, and Engkvist]{inchiguide}
Laurianne David, Amol Thakkar, Roc{\'i}o Mercado, and Ola Engkvist.
\newblock Molecular representations in ai-driven drug discovery: a review and
  practical guide.
\newblock \emph{Journal of Cheminformatics}, 12\penalty0 (1):\penalty0 56, Sep
  2020.
\newblock ISSN 1758-2946.
\newblock \doi{10.1186/s13321-020-00460-5}.
\newblock URL \url{https://doi.org/10.1186/s13321-020-00460-5}.

\bibitem[Dosovitskiy et~al.(2020)Dosovitskiy, Beyer, Kolesnikov, Weissenborn,
  Zhaim, Thomas~Unterthiner, Minderer, Heigold, Gelly, Uszkoreit, and
  Houlsby]{vit}
Alexey Dosovitskiy, Lucas Beyer, Alexander Kolesnikov, Dirk Weissenborn,
  Xiaohua Zhaim, Mostafa~Dehghani Thomas~Unterthiner, Matthias Minderer, Georg
  Heigold, Sylvain Gelly, Jakob Uszkoreit, and Neil Houlsby.
\newblock An image is worth 16x16 words: Transformers for image recognition at
  scale.
\newblock \emph{arXiv:2010.11929}, 2020.

\bibitem[Goh et~al.(2017{\natexlab{a}})Goh, Hodas, Siegel, and
  Vishnu]{smiles2vec}
Garrett Goh, Nathan Hodas, Charles Siegel, and Abhinav Vishnu.
\newblock Smiles2vec: An interpretable general-purpose deep neural network for
  predicting chemical properties, 12 2017{\natexlab{a}}.

\bibitem[Goh et~al.(2017{\natexlab{b}})Goh, Siegel, Vishnu, Hodas, and
  Baker]{chemception}
Garrett Goh, Charles Siegel, Abhinav Vishnu, Nathan Hodas, and Nathan Baker.
\newblock Chemception: A deep neural network with minimal chemistry knowledge
  matches the performance of expert-developed qsar/qspr models, 06
  2017{\natexlab{b}}.

\bibitem[He et~al.(2016)He, Zhang, Ren, and Sun]{resnet}
Kaiming He, Xiangyu Zhang, Shaoqing Ren, and Jian Sun.
\newblock “deep residual learning for image recognition".
\newblock \emph{arXiv:1512.03385}, 2016.

\bibitem[Heller et~al.(2015)Heller, McNaught, Pletnev, Stein, and
  Tchekhovskoi]{inchi}
Stephen~R. Heller, Alan McNaught, Igor Pletnev, Stephen Stein, and Dmitrii
  Tchekhovskoi.
\newblock Inchi, the iupac international chemical identifier.
\newblock \emph{Journal of Cheminformatics}, 7\penalty0 (1):\penalty0 23, May
  2015.
\newblock ISSN 1758-2946.
\newblock \doi{10.1186/s13321-015-0068-4}.
\newblock URL \url{https://doi.org/10.1186/s13321-015-0068-4}.

\bibitem[Hossain et~al.(2019)Hossain, Sohel, Shiratuddin, and
  Laga]{hossain2019comprehensive}
MD~Zakir Hossain, Ferdous Sohel, Mohd~Fairuz Shiratuddin, and Hamid Laga.
\newblock A comprehensive survey of deep learning for image captioning.
\newblock \emph{ACM Computing Surveys (CsUR)}, 51\penalty0 (6):\penalty0 1--36,
  2019.

\bibitem[Mao et~al.(2014)Mao, Xu, Yang, Wang, Huang, and Yuille]{mao2014deep}
Junhua Mao, Wei Xu, Yi~Yang, Jiang Wang, Zhiheng Huang, and Alan Yuille.
\newblock Deep captioning with multimodal recurrent neural networks (m-rnn).
\newblock \emph{arXiv preprint arXiv:1412.6632}, 2014.

\bibitem[Oldenhof et~al.(2020)Oldenhof, Arany, Moreau, and Simm]{chemgrapher}
Martijn Oldenhof, Adam Arany, Yves Moreau, and Jaak Simm.
\newblock Chemgrapher: Optical graph recognition of chemical compounds by deep
  learning.
\newblock \emph{Journal of Chemical Information and Modeling}, 60\penalty0
  (10):\penalty0 4506--4517, 2020.
\newblock \doi{10.1021/acs.jcim.0c00459}.
\newblock URL \url{https://doi.org/10.1021/acs.jcim.0c00459}.
\newblock PMID: 32924466.

\bibitem[Park et~al.(2009)Park, Rosania, Shedden, Nguyen, Lyu, and
  Saitou]{park2009automated}
Jungkap Park, Gus~R Rosania, Kerby~A Shedden, Mandee Nguyen, Naesung Lyu, and
  Kazuhiro Saitou.
\newblock Automated extraction of chemical structure information from digital
  raster images.
\newblock \emph{Chemistry Central Journal}, 3\penalty0 (1):\penalty0 1--16,
  2009.

\bibitem[Park et~al.(2010)Park, Saitou, and Rosania]{rule1}
Jungkap Park, Kazuhiro Saitou, and Gus Rosania.
\newblock Image-based automated chemical database annotation with ensemble of
  machine-vision classifiers.
\newblock In \emph{2010 IEEE International Conference on Automation Science and
  Engineering}, pages 168--173, 2010.
\newblock \doi{10.1109/COASE.2010.5584695}.

\bibitem[Paul et~al.(2018)Paul, Jha, Al-Bahrani, Liao, Choudhary, and
  Agrawal]{chemixnet}
Arindam Paul, Dipendra Jha, Reda Al-Bahrani, Wei-Keng Liao, Alok Choudhary, and
  Ankit Agrawal.
\newblock Chemixnet: Mixed dnn architectures for predicting chemical properties
  using multiple molecular representations, 11 2018.

\bibitem[Rajan et~al.(2020)Rajan, Zielesny, and Steinbeck]{decimer}
Kohulan Rajan, Achim Zielesny, and Christoph Steinbeck.
\newblock Decimer - towards deep learning for chemical image recognition, 06
  2020.

\bibitem[Rajan et~al.(2021)Rajan, Brinkhaus, Sorokina, Zielesny, and
  Steinbeck]{segment}
Kohulan Rajan, Henning~Otto Brinkhaus, Maria Sorokina, Achim Zielesny, and
  Christoph Steinbeck.
\newblock Decimer-segmentation: Automated extraction of chemical structure
  depictions from scientific literature.
\newblock \emph{Journal of Cheminformatics}, 13\penalty0 (1):\penalty0 20, Mar
  2021.
\newblock ISSN 1758-2946.
\newblock \doi{10.1186/s13321-021-00496-1}.
\newblock URL \url{https://doi.org/10.1186/s13321-021-00496-1}.

\bibitem[Ren et~al.(2016)Ren, He, Girshick, and Sun]{fasterrcnn}
Shaoqing Ren, Kaiming He, Ross Girshick, and Jian Sun.
\newblock “faster r-cnn: Towards real-time object detection with region
  proposal networks".
\newblock \emph{arXiv:1506.01497}, 2016.

\bibitem[Rennie et~al.(2017{\natexlab{a}})Rennie, Marcheret, Mroueh, Ross, and
  Goel]{rennie2017self}
Steven~J Rennie, Etienne Marcheret, Youssef Mroueh, Jerret Ross, and Vaibhava
  Goel.
\newblock Self-critical sequence training for image captioning.
\newblock In \emph{Proceedings of the IEEE Conference on Computer Vision and
  Pattern Recognition}, pages 7008--7024, 2017{\natexlab{a}}.

\bibitem[Rennie et~al.(2017{\natexlab{b}})Rennie, Marcheret, Mroueh, Ross, and
  Goel]{train}
Steven~J Rennie, Etienne Marcheret, Youssef Mroueh, Jerret Ross, and Vaibhava
  Goel.
\newblock “self-critical sequence training for image captioning".
\newblock \emph{arXiv:1612.00563}, 2017{\natexlab{b}}.

\bibitem[Sadawi et~al.(2012)Sadawi, Sexton, and Sorge]{sadawi2012chemical}
Noureddin~M Sadawi, Alan~P Sexton, and Volker Sorge.
\newblock Chemical structure recognition: a rule-based approach.
\newblock In \emph{Document Recognition and Retrieval XIX}, volume 8297, page
  82970E. International Society for Optics and Photonics, 2012.

\bibitem[Staker et~al.(2019)Staker, Marshall, Abel, and
  McQuaw]{staker2019molecular}
Joshua Staker, Kyle Marshall, Robert Abel, and Carolyn~M McQuaw.
\newblock Molecular structure extraction from documents using deep learning.
\newblock \emph{Journal of chemical information and modeling}, 59\penalty0
  (3):\penalty0 1017--1029, 2019.

\bibitem[Vaswani et~al.(2017)Vaswani, Shazeer, Parmar, Uszkoreit, Jones, Gomez,
  Kaiser, and Polosukhin]{attention}
Ashish Vaswani, Noam Shazeer, Niki Parmar, Jakob Uszkoreit, Llion Jones,
  Aidan~N Gomez, Lukasz Kaiser, and Illia Polosukhin.
\newblock “attention is all you need".
\newblock \emph{arXiv:1706.03762}, 2017.

\bibitem[Vinyals et~al.(2016)Vinyals, Toshev, Bengio, and
  Erhan]{vinyals2016show}
Oriol Vinyals, Alexander Toshev, Samy Bengio, and Dumitru Erhan.
\newblock Show and tell: Lessons learned from the 2015 mscoco image captioning
  challenge.
\newblock \emph{IEEE transactions on pattern analysis and machine
  intelligence}, 39\penalty0 (4):\penalty0 652--663, 2016.

\bibitem[Wiswesser(1968)]{19cen}
William~J. Wiswesser.
\newblock 107 years of line-formula notations (1861-1968).
\newblock \emph{Journal of Chemical Documentation}, 8\penalty0 (3):\penalty0
  146--150, 1968.
\newblock \doi{10.1021/c160030a007}.
\newblock URL \url{https://doi.org/10.1021/c160030a007}.

\bibitem[Xu et~al.(2015)Xu, Ba, Kiros, Cho, Courville, Salakhudinov, Zemel, and
  Bengio]{xu2015show}
Kelvin Xu, Jimmy Ba, Ryan Kiros, Kyunghyun Cho, Aaron Courville, Ruslan
  Salakhudinov, Rich Zemel, and Yoshua Bengio.
\newblock Show, attend and tell: Neural image caption generation with visual
  attention.
\newblock In \emph{International conference on machine learning}, pages
  2048--2057. PMLR, 2015.

\bibitem[You et~al.(2019)You, Li, Hseu, Song, Demmel, and Hsieh]{LAMB}
Yang You, Jing Li, Jonathan Hseu, Xiaodan Song, James Demmel, and Cho{-}Jui
  Hsieh.
\newblock Reducing {BERT} pre-training time from 3 days to 76 minutes.
\newblock \emph{CoRR}, abs/1904.00962, 2019.
\newblock URL \url{http://arxiv.org/abs/1904.00962}.

\bibitem[Zheng et~al.(2020)Zheng, Lu, Zhao, Zhu, Luo, Wang, Fu, Feng, Xiang,
  and Torr]{imageseg}
Sixiao Zheng, Jiachen Lu, Hengshuang Zhao, Xiatian Zhu, Zekun Luo, Yabiao Wang,
  Yanwei Fu, Jianfeng Feng, Tao Xiang, and Philip~HS Torr.
\newblock “rethinking semantic segmentation from a sequence-tosequence
  perspective with transformers.
\newblock \emph{arXiv preprint arXiv:2012.15840}, 2020.

\end{thebibliography}

\end{document}